%% file: acl_latex.tex
\pdfoutput=1

\documentclass[11pt]{article}

\usepackage{acl}

\usepackage{times}
\usepackage{latexsym}

\usepackage[T1]{fontenc}

\usepackage[utf8]{inputenc}

\usepackage{microtype}

\usepackage{inconsolata}

\usepackage{graphicx}

\usepackage{booktabs} 
\usepackage{arydshln} 
\usepackage{xcolor}   
\usepackage{CJKutf8}
\usepackage{multirow}
\usepackage{comment}
\usepackage{float}
\usepackage{ulem}
\usepackage{amsmath}
\usepackage[most]{tcolorbox}
\title{Investigating and Scaling up Code-Switching for \\
Multilingual Language Model Pre-Training}
\author{
  \textbf{Zhijun Wang\textsuperscript{1,2}},
  \textbf{Jiahuan Li\textsuperscript{2}},
  \textbf{Hao Zhou\textsuperscript{1}},
  \textbf{Rongxiang Weng\textsuperscript{2}},
  \textbf{Jingang Wang\textsuperscript{2}},
\\
  \textbf{Xin Huang\textsuperscript{3}},
  \textbf{Xue Han\textsuperscript{3}},
  \textbf{Junlan Feng\textsuperscript{3}},
  \textbf{Chao Deng\textsuperscript{3}},
  \textbf{Shujian Huang\textsuperscript{1}}\thanks{The Corresponding author.}
\\
  \textsuperscript{1}National Key Laboratory for Novel Software Technology, Nanjing University, China\\
  \textsuperscript{2}Meituan Inc.
  \textsuperscript{3}China Mobile Research Beijing, China
\\
    {\{wangzj,zhouh\}@smail.nju.edu.cn,\{huangsj\}@nju.edu.cn}\\
    {\{lijiahuan04,wengrongxiang,wangjingang02\}@meituan.com}\\
    {\{huangxinyjy,hanxueai,fengjunlan,dengchao\}@chinamobile.com}
}

\begin{document}
\maketitle
\begin{CJK*}{UTF8}{gkai}
\begin{abstract}
Large language models (LLMs) exhibit remarkable multilingual capabilities despite the extreme language imbalance in the pre-training data. In this paper, we closely examine the reasons behind this phenomenon, focusing on the pre-training corpus. We find that the existence of code-switching, alternating between different languages within a context, is key to multilingual capabilities.
We conduct an analysis to investigate code-switching in the pre-training corpus, examining its presence and categorizing it into four types within two quadrants. We then assess its impact on multilingual performance. These types of code-switching data are unbalanced in proportions and demonstrate different effects on facilitating language transfer.
To better explore the power of code-switching for language alignment during pre-training, we investigate the strategy of synthetic code-switching. We continuously scale up the synthetic code-switching data and observe remarkable improvements in both benchmarks and representation space. Extensive experiments indicate that incorporating synthetic code-switching data enables better language alignment and generalizes well to high, medium, and low-resource languages with pre-training corpora of varying qualities. Code and scripts are freely available at \url{https://github.com/NJUNLP/SynCS}.
\end{abstract}

\input{introduction}

\input{understanding}

\input{scaling}
\input{realated_work}
\input{conclusion}
\input{limitation}


\bibliography{custom}

\input{appendix}
\end{CJK*}
\end{document}

%% file: introduction.tex
\section{Introduction}
Large Language Models (LLMs) such as ChatGPT~\citep{chatgpt}, GPT-4~\citep{achiam2023gpt}, Llama2~\citep{touvron2023llama}, Llama3~\citep{dubey2024llama}, and Qwen2.5~\citep{yang2024qwen2} have demonstrated remarkable performance across various tasks, including multiple-choice question-answering~\citep{robinson2023leveraging}, summarization~\citep{pu2023summarization}, and reasoning~\citep{yu2023metamath}. Meanwhile, LLMs also demonstrate excellent multilingual capabilities.
Among them, some models are pre-trained on corpora not specifically designed for multilingual use~\citep{touvron2023llama}, while others are pre-trained on corpora containing only a small fraction of multilingual data~\citep{dubey2024llama}.
Despite the extreme language imbalance in the pre-training corpus~\citep{ranta2021linguistic}, LLMs demonstrate impressive cross-lingual transfer to some extend~\citep{pires-etal-2019-multilingual,kargaran2024mexa}. This raises the question: where do these cross-lingual transfers come from?
\begin{figure}[t]
\centering
\includegraphics[width=0.41\textwidth]{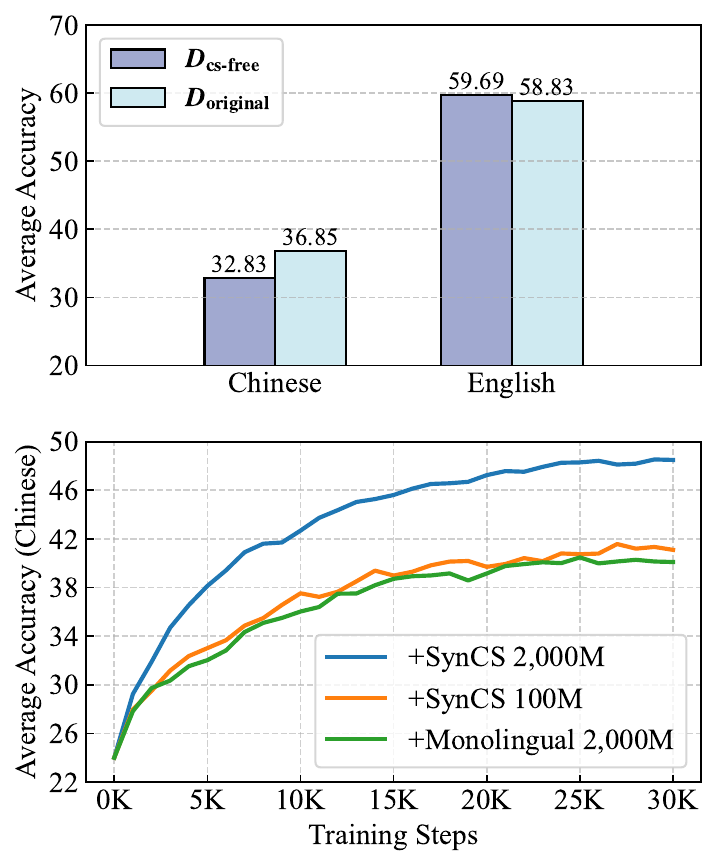}
\caption{Performance of models pre-trained on language-imbalance data (60B En and 600M Zh, 100:1). $\mathcal{D}_\text{cs-free}$ in the upper sub-graph means the natural code-switching is removed. ``+SynCS'' and ``+Monolingual'' in the lower sub-graph denote adding synthetic code-switching data and monolingual data, respectively. The numbers represent newly added Zh tokens.}
\label{figure:clean}
\end{figure}

\begin{table*}[t]
\centering
\footnotesize
\setlength{\tabcolsep}{3pt} 
{
\begin{tabular}{p{0.1\textwidth}p{0.875\textwidth}} 
\toprule
{\bf Category} & {\bf Example} \\
\midrule
Sent-Annt. & Now depending on where you shop in China, sometimes you need to bargain for what you are buying. \uline{Mike, the fruits stand is just ahead, let’s buy some fruit OK?} (\textcolor{blue}{\small{麦克, 前面有一个水果摊, 我们买点儿水果吧.}}) \\
\midrule
Sent-Repl. & Can you name some traditional Chinese festivals? Do you like them? Why? \textcolor{blue}{\small{这道题的目的是要求考生陈述出来传统文化的重要性。}} Traditional cultures should be protected. because first...... [The Chinese sentence means ``The purpose of this question is to require candidates to state the importance of traditional culture.'']\\
\midrule
Token-Annt. & The customs of the spring festival: 1. \underline{Putting up Spring Couplet} (\textcolor{blue}{\small{贴春联}}) and \underline{Burning Firecrackers} (\textcolor{blue}{\small{放鞭炮}}). \\
\midrule
Token-Repl. & You can use the above picture and add some related words, such as \textcolor{blue}{\small{剃须刀、字典、镜子、毛巾、冰箱、微波炉、电脑}} and \textcolor{blue}{\small{书橱}}. Classify these words and fill in the table. [These Chinese words mean ``razor'', ``dictionary'', ``mirror'', ``towel'', ``refrigerator'', ``microwave'', ``computer'', and ``bookcase'', respectively.] \\
\bottomrule
\end{tabular}
}
\caption{Examples of code-switching types in FineWeb-Edu. For annotation types, annotations are typically placed in parentheses, with the annotated text underlined. For replacement types, code-switching occurs within the original text, and explanations are appended in brackets after the example.}
\label{tab:example}
\end{table*}
Code-switching, also known as code-mixing or language alternation, is the process of alternating between two or more languages in a single conversation~\citep{8554413}. This type of data puts concepts from different languages within the same context, creating favorable conditions for potential language transfer learning in LLMs.
Many works attempt to leverage code-switching on multilingual tasks. ~\citet{yoo2024code,li-etal-2024-prealign} reveal the effects of synthetic code-switching data in cross-lingual transfers. ~\citet{briakou-etal-2023-searching} investigate the incidental bilingualism in the unreasonable translation capabilities of LLMs. However, there is a lack of detailed analysis of code-switching in multilingual pre-training.

To investigate the effects of code-switching, we pre-train a 1.5B model on 60B tokens with extreme language imbalance (100:1). Taking English and Chinese as the high and low-resource language examples, we initially explore the natural code-switching phenomenon of two high-quality pre-training corpora. We conduct a model-based method to analyze and categorize four common code-switching types. Subsequently, we conduct experiments to assess the impact of various code-switching on cross-lingual transfer.

Building on this analysis, we propose to enhance the advantages of code-switching by incorporating synthetic code-switching data during pre-training, valued for its controllability and flexibility. Through a series of scaling experiments, synthetic code-switching (SynCS) significantly improves cross-lingual transfer, outperforming the addition of 20 times the amount of monolingual data with natural code-switching. Further analysis shows that models trained on SynCS data obtain improved multilingual alignment in the representation space. Finally, we expand our approach to multilingual settings, encompassing high, medium, and low-resource languages, showcasing the generalization of SynCS across languages.

\noindent In summary, our findings are:
\begin{itemize}
\item Natural Code-Switching in Pre-Training Data: In FineWeb-Edu~\citep{NEURIPS2024_370df50c}, 0.4\% of documents contain English-Chinese code-switching, compared to 51.6\% in Chinese-FineWeb-Edu-v2~\citep{yu2025opencsg}. These instances, categorized into four types, enhance multilingual transfer despite their imbalance.

\item Role of Natural Code-Switching: Natural code-switching plays a crucial role in facilitating cross-lingual transfer. As illustrated in Figure~\ref{figure:clean}, models trained without it experience a notable performance drop.

\item We introduce SynCS, a flexible framework for synthesizing code-switching with precise control over density and magnitude. Models trained with SynCS exhibit superior multilingual alignment, surpassing the performance achieved by adding 20x monolingual data, as shown in Figure~\ref{figure:clean}.

\end{itemize}

%% file: understanding.tex
\section{Measuring Code-Switching}
\subsection{Categorizing Code-Switching}
Based on our empirical analysis, code-switching segments are categorized into \textbf{Sentence-Level} and \textbf{Token-Level}, each further divided into \textbf{Annotation} and \textbf{Replacement}. Considering languages A and B, the code-switching types are defined as follows:
\begin{itemize}\label{typing}
\item \textbf{Sentence-Level-Annotation} (denoted as \textbf{Sent-Annt.}): In a continuous sequence of sentences in the context of language A, some sentences are annotated by their translation in language B, commonly appearing in parentheses. The semantics represented by these sentences appear in both languages A and B.

\begin{figure}[t]
\centering
\includegraphics[width=0.46\textwidth]{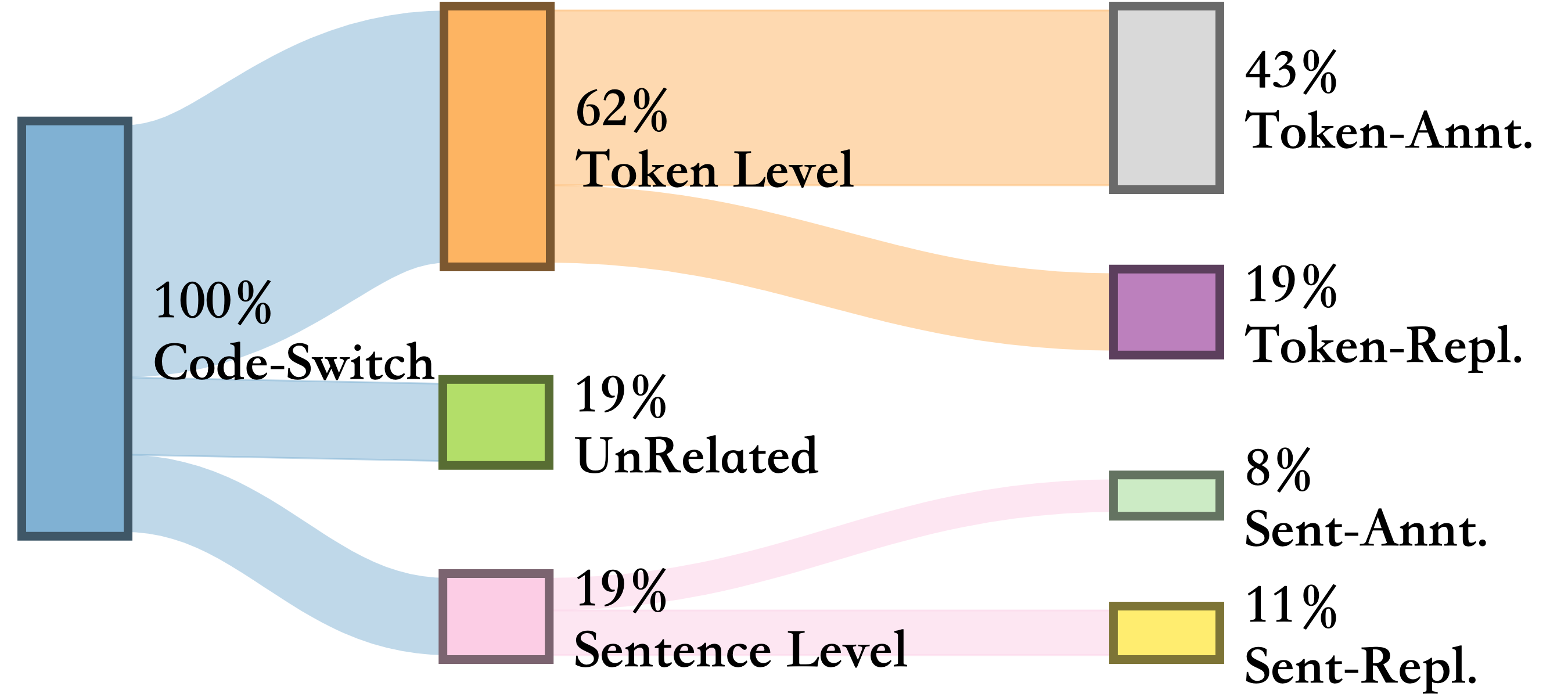}
\caption{Distribution of different types of En->Zh code-switching in FineWeb-Edu.}
\label{figure:endata}
\end{figure}

\item \textbf{Sentence-Level-Replacement} (denoted as \textbf{Sent-Repl.}): In a continuous sequence of sentences in the context of language A, some sentences are replaced by their translation in language B. The semantics represented by these sentences appear only in language B.

\item \textbf{Token-Level-Annotation} (denoted as \textbf{Token-Annt.}): In a sentence of language A, some tokens are annotated by their translation in language B, commonly appearing in parentheses. The concepts represented by these tokens appear in both languages A and B.

\item \textbf{Token-Level-Replacement} (denoted as \textbf{Token-Repl.}): In a sentence of language A, some tokens are replaced by their translation in language B. The concepts represented by these tokens appear only in language B.

\end{itemize}
Table~\ref{tab:example} presents examples for each type of Chinese code-switching in English data.
In our following discussions, ``Code-Switching in A'' refers to containing text of B in the context of A. 

\subsection{Detecting Code-Switching Segments} 
To investigate the characteristics of natural code-switching, we need first detect all code-switching segments.
Code-switching in both high-resource and low-resource languages can enhance cross-lingual transfer, so we begin by identifying code-switching documents in each language dataset. These documents are segmented into sentences, with each sentence tagged by its language. Sentences entirely in one language, differing from the document's language, are classified as sentence-level code-switching, while sentences incorporating both languages are considered token-level code-switching.

The strategy for classifying segments into Annt. and Repl. differs between sentence-level and token-level code-switching. For sentence level, this process is indeed identifying translation pairs. We employ a cross-lingual encoder to find semantically aligned sentence pairs, following \citet{briakou-etal-2023-searching}. For token level, we leverage SOTA LLMs to classify. Additionally, we use LLMs to detect unrelated code-switching segments, which may result from nonsensical content or language recognition errors (such as text of Japanese).

To simplify our analysis, we choose to explore the Chinese and English code-switching data which has completely different scripts. We choose two high-quality corpora: FineWeb-Edu~\citep{NEURIPS2024_370df50c} and Chinese-FineWeb-Edu-v2~\citep{yu2025opencsg}. More details are illustrated in section~\ref{detecting}.

\subsection{Counting Code-Switching Segments}

\begin{figure}[t]
\centering
\includegraphics[width=0.46\textwidth]{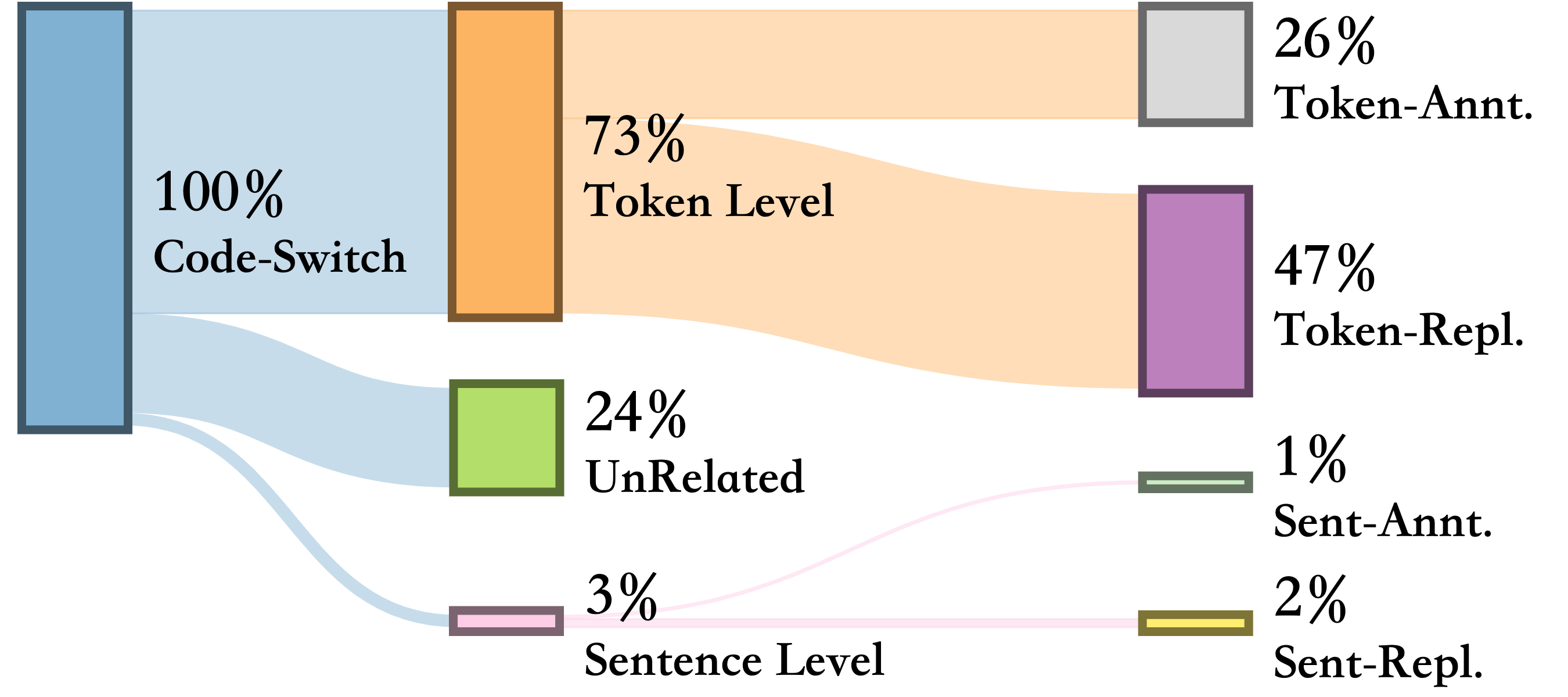}
\caption{Distribution of different types of Zh->En code-switching in Chinese-FineWeb-Edu-v2.}
\label{figure:zhdata}
\end{figure}

We calculate the ratio of different code-switching types at the segment granularity.

In FineWeb-Edu, 0.4\% of documents contain Chinese-English code-switching. 
Figure~\ref{figure:endata} shows the distribution of different types. 
19\% code-switching segments fall under unrelated, most of which are segments containing characters of Japanese or nonsense text.
In the remaining 81\% code-switching documents, the main type is token-level (62\%), among which Annt. accounts the most (43\%).
For sentence-level code-switching, the proportion of Annt. and Repl. are similar. Examples of each type are illustrated in Table~\ref{tab:example} and section~\ref{En-side-natural}.

In Chinese-FineWeb-Edu-v2, 51.2\% of documents contain Chinese-English code-switching. 
Figure~\ref{figure:zhdata} demonstrates the distribution. 
24\% are unrelated. 
The proportion of sentence-level code-switching is very small, approximately 3\%, with 1\% being Annt. and the rest 2\% being Repl. In contrast to FineWeb-Edu, the Token-Repl. code-switching accounts more than the Token-Annt. code-switching.
This is caused by the frequent use of proper noun, such as ``Microsoft'', ``CAR-T'' (Chimeric Antigen Receptor T-Cell) and so on.
Examples of each type are illustrated in section~\ref{zh-side-natural}.

\begin{table}[t]
\centering
\footnotesize
\scalebox{1}{
\begin{tabular}{lccccc}
\toprule
{\bf Data} &{\bf PPL \(\downarrow\)} &{\bf MEXA} &{\bf Acc. Avg.} \\
\midrule
$\mathcal{D}_\text{original}$ &41.2 &0.66 &36.9 \\
$\mathcal{D}_\text{control}$ &40.5 &0.66 &37.9 \\
$\mathcal{D}_\text{cs-free}$ &66.0 &0.43 &32.8 \\
\bottomrule
\end{tabular}}
\caption{Comparison of Chinese performance of models trained on different data. ``Acc. Avg.'' is the average accuracy on Hellaswag and ARC-Easy.}
\label{tab:clean}
\end{table}

\section{Analyzing the Impact of Code-Switching}
Based on our analysis of natural code-switching, we investigate its impact on cross-lingual transfer.

\subsection{Experiment Setup}
\paragraph{Pre-Training Recipes}
We sample 60B English tokens from FineWeb-Edu and 600M Chinese tokens from Chinese-FineWeb-Edu-v2 to simulate the language imbalance (100:1) pre-training\footnote{We follow~\citet{li-etal-2024-prealign}'s language imbalance pre-training settings.}. 
A 1.5B model is trained from scratch on this sampled data to explore the cross-lingual transfer during pre-training.
The hyper-parameters for pre-training are detailed in section~\ref{pretrain}.
\paragraph{Evaluation Recipes}
We use the perplexity on Wikipedia~\citep{wikidump}, and the accuracy on Hellaswag~\citep{zellers-etal-2019-hellaswag} and ARC-Easy~\citep{clark2018think} to evaluate the performance in each language.
Besides, we present MEXA~\citep{kargaran2024mexa} scores, which assess alignment between English and non-English languages using parallel sentences to evaluate language transfer.
More evaluation details are illustrated in Section~\ref{eval}.

\subsection{Ablating All Code-Switching}
We employ a document-substitute-based ablating method. Let $\mathcal{M}$ denote the documents used for pre-training and $\mathcal{P}$ denote the homologous holdout documents, where the partitions are defined as:
$$
\mathcal{M} = \mathcal{M}_{\text{wcs}} \cup \mathcal{M}_{\text{wocs}}, \quad 
\mathcal{P} = \mathcal{P}_{\text{wcs}} \cup \mathcal{P}_{\text{wocs}}
$$
where "wcs" and "wocs" refer to documents with and without code-switching respectively.

To investigate the overall impact of code-switching, we construct the code-switching-free dataset $\mathcal{D}_\text{cs-free}$ through document substitution:
$$
\mathcal{D}_\text{cs-free} = \mathcal{M}_\text{wocs} \cup \mathcal{S}
$$
where $\mathcal{S}$ is a randomly sampled subset from $\mathcal{P}_{\text{wocs}}$ satisfying $\mathcal{S} \subseteq \mathcal{P}_{\text{wocs}}$ and $|\mathcal{S}| = |\mathcal{M}_{\text{wcs}}|$ to maintain equivalent corpus size.

To control for potential confounding factors from newly introduced documents $\mathcal{S}$, we further build a control dataset $\mathcal{D}_\text{control}$:
$$
\mathcal{D}_\text{control} = (\mathcal{M} \setminus \mathcal{T}) \cup \mathcal{S}
$$
where $\mathcal{T}$ is a randomly sampled subset from $\mathcal{M}_{\text{wocs}}$ with $|\mathcal{T}| = |\mathcal{S}|$ to maintain equivalent corpus size.

\begin{figure}[t]
\centering
\includegraphics[width=0.485\textwidth]{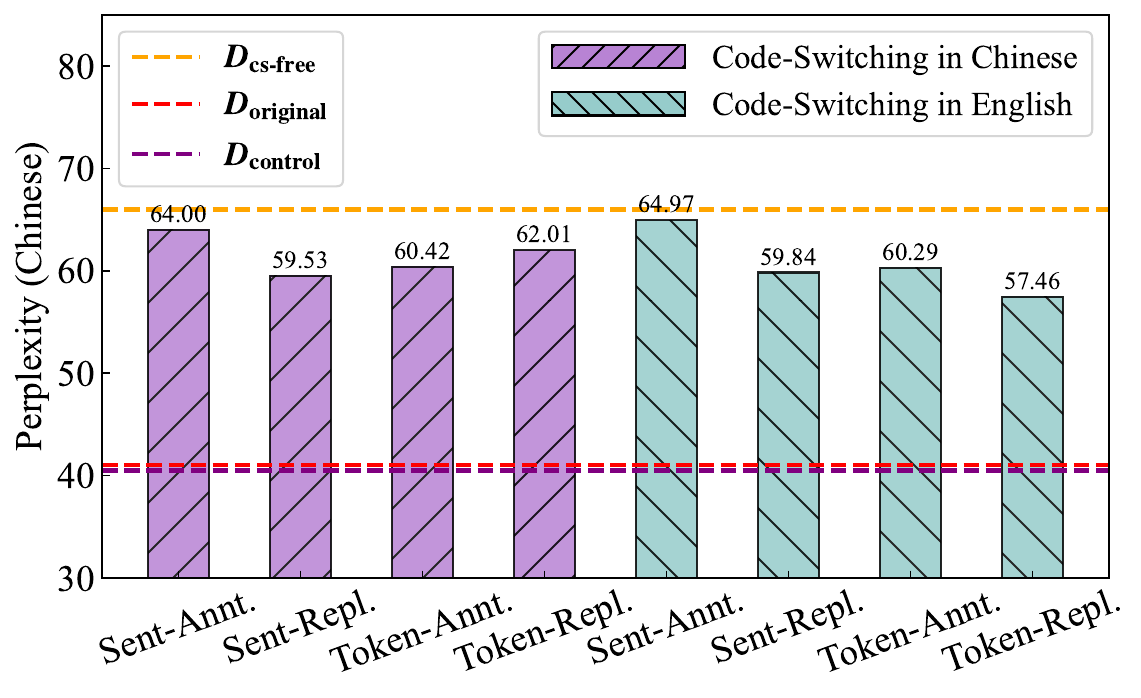}
\caption{Impact of different types of natural code-switching on the cross-lingual transfer.}
\label{figure:ablating}
\end{figure}

\paragraph{Natural Code-Switching Plays a Crucial Role in Cross-Lingual Transfer}
In Table~\ref{tab:clean}, the perplexity of the model trained on $\mathcal{D}_\text{cs-free}$ shows a significant increase compared to that of $\mathcal{D}_\text{control}$ (40.5 to 66.0), and the benchmark performance also decreases by about 5 points. Without natural code-switching, the MEXA alignment score of the model drops significantly (0.66 to 0.43), indicating a worse multilingual alignment in hidden states. These results reveal the importance of natural code-switching in cross-lingual transfer.

\subsection{Ablating Individual Type}
To further investigate the impact of code-switching in various formats, we conduct experiments trained on data containing only one type. 
Since the ablation for each type shows an imperceptible difference in benchmarks, we mainly report the perplexity.
Figure~\ref{figure:ablating} demonstrates the results. 

\paragraph{Less Tokens but Better Transfer}
For Repl. code-switching in Chinese, the number of tokens in Chinese is actually decreasing from the original 600M since some tokens are replaced by its translation. 
However, leveraging Repl. code-switching can still reduce the perplexity, indicating the potential cross-lingual transfer.
Sent-Repl. presents the best effects on cross-lingual transfer, even though it only accounts for 2\%.

\paragraph{Repl. Contributes More than Annt.}
\label{replbetter}
For code-switching in English, Repl. demonstrates better effects than Annt., as shown in Figure~\ref{figure:ablating}.
We suppose that while the concepts represented by code-switched tokens appear twice in both languages in Annt., the model may pay less attention to the Chinese tokens during training. This process may degrade the potential transfer learning.

\paragraph{Translation Fails in Enhancing Multilingual Transfer}
It is worth noting that Sent-Annt. in both English and Chinese, show the worst effects compared to other types. This suggests that while parallel sentences in the pre-training corpus are crucial for the model's translation capabilities~\citep{briakou-etal-2023-searching}, they may not significantly enhance multilingual transfer.

%% file: scaling.tex
\section{Scaling up Code-Switching}
Despite the effectiveness evidenced in the experiment of previous section, the natural code-switching phenomenon is rare and usually restricted to specific domains. In this section, we explore improving multilingual pre-training by synthesizing large-scale documents with code-switching. This method is more flexible and controllable, allowing us to inject code-switching into any document at any density and in any format.

\subsection{Code-Switching Synthesis Pipeline}
Given a collection of documents, we first split them into sentences and randomly select sentences to apply different types of code-switching.

\paragraph{Synthesizing Sentence-level Code-switching} 
For sentence-level code-switching, we use TowerInstruct~\citep{colombo2024tower} to translate each selected sentence. 
When conducting Sent-Repl., the source sentence is directly replaced with its translation.
When conducting Sent-Annt., the source sentence is preserved with its translation following behind in parentheses, which is the most frequent pattern for natural Sent-Annt.

\paragraph{Synthesizing Token-level Code-Switching} 
Currently, there is a lack of flexible and low-cost methods for synthesizing high-quality token-level code-switching.
~\citet{li-etal-2024-prealign} conduct rule-based method using a bilingual dictionary. However, it suffers from the one-to-many problem of word alignment and fails to select suitable tokens to replace or annotate.
~\citet{yoo2024code} leverages GPT-4o and parallel sentences to synthesize high quality Token-Repl. code-switching data. However, it is expensive when scaling up and can not be used on monolingual documents.
Empirically, we also find that SOTA LLMs struggle to generate token-level code-switching content given only monolingual text.

To synthesize high-quality token-level code-switching without requiring parallel sentences at a low-cost, we introduce a data-based distillation method. Initially, inspired by~\citet{yoo2024code}, we leverage GPT-4o-mini to generate high-quality Token-Annt. and Token-Repl. code-switching data based on parallel sentences.
Then we construct Supervised Fine-Tuning (SFT) data by only preserving the sentence of one language in the instruction, resulting in a multilingual dataset.
A small language model is then fine-tuned on this dataset, learning to synthesize token-level code-switching.
Practically, we select Qwen2.5-3B-Instruct as the base model, taking both speed and effect into consideration.
The resultant model can rapidly generate diverse and high-quality code-switching data at a low cost.
The prompts for generating SFT data and fine-tuning are illustrated in section~\ref{prompts_syncs}.

\subsection{Scaling up Code-Switching in Chinese}
\begin{figure}[t]
\centering
\includegraphics[width=0.42\textwidth]{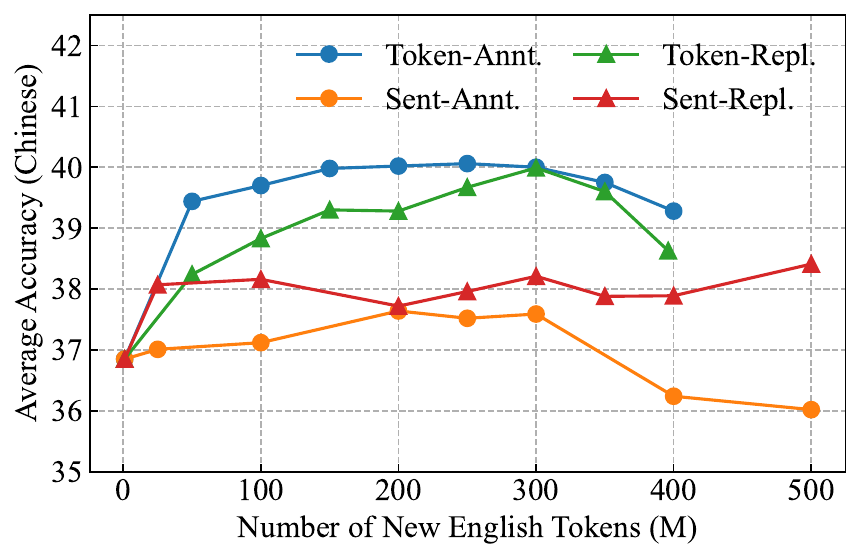}
\caption{Scaling code-switching in Chinese: Average accuracy on Hellaswag and ARC-Easy in Chinese.}
\label{figure:zhscale}
\end{figure}

To assess whether scaling on the low-resource language enhances cross-lingual transfer, we modify the 600M Chinese documents to include English code-switching segments. In Figure~\ref{figure:zhscale}, we increase the number of newly added English tokens from 0M to 500M by adjusting the ratio of modified sentences.

\paragraph{Improved Cross-Lingual Transfer with Code-Switching scaling in Chinese}
As we modify more sentences from the 600M Chinese documents, the performance in Chinese continues to improve. Adding 300M new English tokens results in significant improvements (39.99 vs 36.85). This demonstrates that SynCS in the Chinese effectively enhances cross-lingual transfer.

\paragraph{The Importance of Chinese Monolingual Data}
Beyond 300M, all four types of code-switching in Chinese exhibit a notable performance drop. This decline is due to excessive alterations of the original Chinese documents as we modify over 60\% of the sentences. This highlights the importance of retaining the low-resource language monolingual data. Notably, even with 100\% modification, Token-Annt. still presents substantial improvements (+2.43). 

\paragraph{Token-Level Code-Switching Exceeds Sentence-Level}
In Figure~\ref{figure:zhscale}, Token-Annt. and Token-Repl. consistently exceeds Sent-Annt. and Sent-Repl., with a maximum gap of 1.58 points. The scalability of sentence-level code-switching in Chinese appears to be limited, suggesting that token-level code-switching is more suitable for the low-resource language.

\subsection{Scaling up Code-Switching in English}
\begin{figure}[t]
\centering
\includegraphics[width=0.42\textwidth]{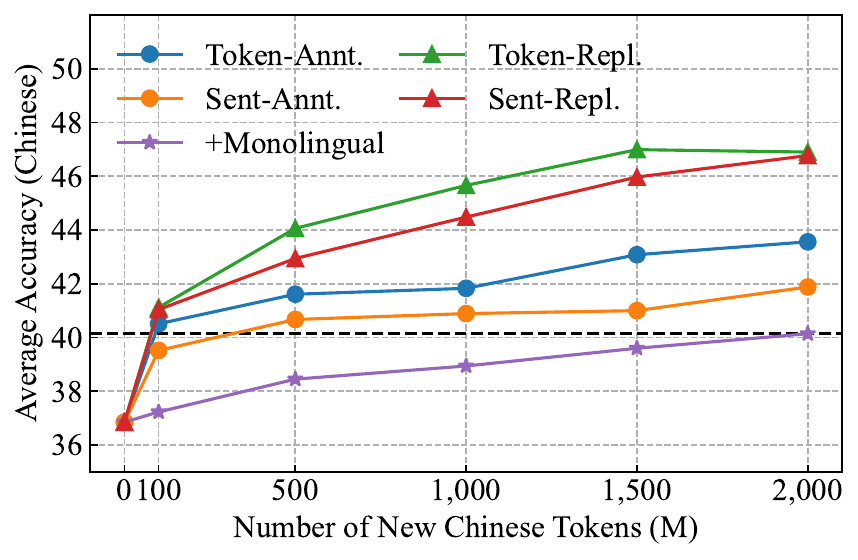}
\caption{Scaling code-switching in English: Average accuracy on Hellaswag and ARC-Easy in Chinese.}
\label{figure:enscale}
\end{figure}

Since code-switching in English increases the token count of Chinese, we explore whether it exhibits better scalability.
We modify only 20\% of the documents (12B) to ensure stable English learning. In Figure~\ref{figure:enscale}, we increase the number of newly added Chinese tokens from 0M to 2,000M by adjusting the ratio of modified sentences.

\paragraph{Greater Efficiency of Code-Switching in English}
The results show the advantages of code-switching in English compared to Chinese. By adding 100M new tokens, the performance of code-switching in English exceeds that of Chinese by 1.42 points. This gap increases with over 100M tokens, reaching a maximum of 6.93 points. As English dominates during pre-training, it allows for extensive code-switching scaling without reducing low-resource langauge tokens.

\paragraph{Superior Scalability of Code-Switching in English}
By scaling the newly added Chinese tokens from 0M to 2,000M, SynCS demonstrates improvements from 0 to 10.14.
This showcases its superior scalability.
In experiments comparing the addition of an equivalent amount of Chinese monolingual tokens from holdout documents, SynCS consistently demonstrates superior performance. At 100 M, SynCS matches or surpasses the performance achieved by adding 20x monolingual data at 2,000M, highlighting its remarkable efficiency.

\begin{figure*}[h]
\centering
\includegraphics[width=0.8\textwidth]{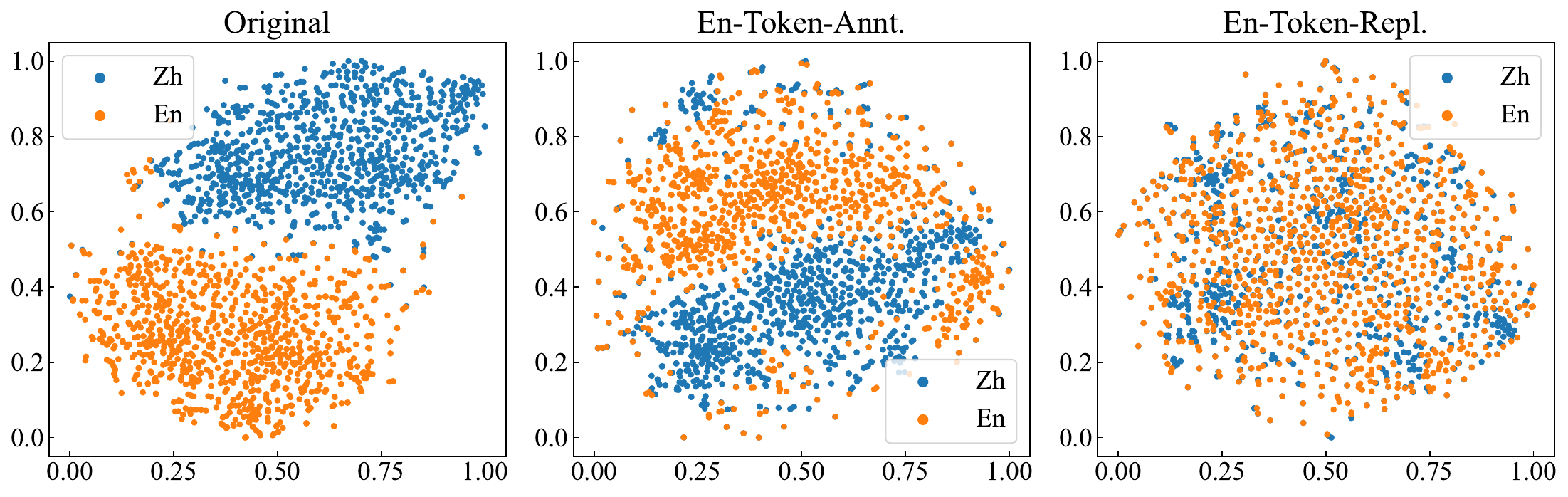}
\caption{T-SNE visualization of parallel sentences' middle-layer hidden states shows significant changes only in En-Token-Repl. and En-Sent-Repl, as illustrated in Figures~\ref{figure:all_tsne}. We take En-Token-Annt. and En-Token-Repl. as examples here.}
\label{figure:tsne}
\end{figure*}

\paragraph{Replacement Transfers Better than Annotation}
Figure~\ref{figure:enscale} shows that Sent-Repl. and Token-Repl. outperform Sent-Annt. and Token-Annt. with faster performance improvements. This is consistent with the ablation study of natural code-switching in section~\ref{replbetter}, which indicates that Repl. in English enhances multilingual performance more than Annt. Figure~\ref{figure:tsne} presents the t-SNE visualizations~\citep{van2008visualizing} of parallel sentences' middle-layer hidden states for models trained on SynCS data of different types. Notably, only Token-Repl. and Sent-Repl. exhibit significant changes, suggesting a more comprehensive cross-lingual transfer process through evenly mixed representations of parallel sentences.

\begin{figure}[t]
\centering
\includegraphics[width=0.45\textwidth]{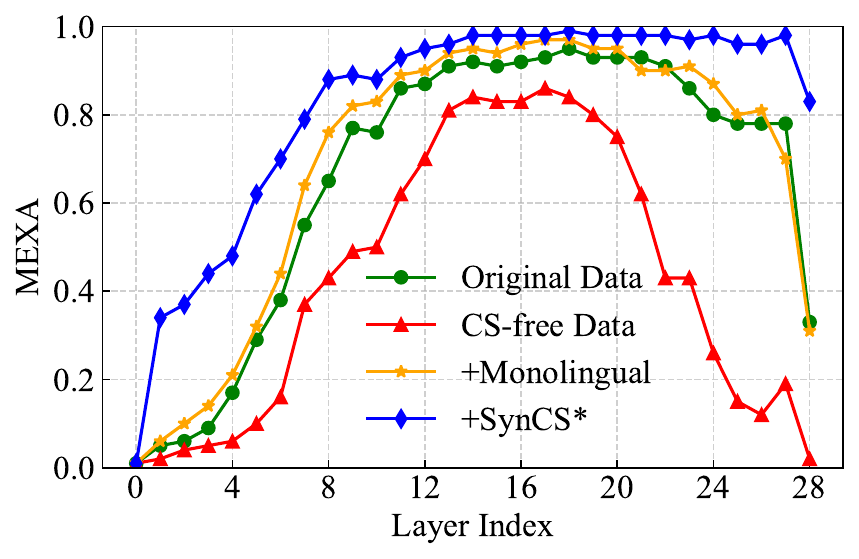}
\caption{The MEXA alignment score comparison.}
\label{figure:mexa}
\end{figure}

\begin{table*}[ht]
\centering
\scriptsize
\scalebox{1}{
\begin{tabular}{lcccccccccc}
\toprule
\multirow{2}{*}{\bf Data} & \multirow{2}{*}{\bf \# New Tokens} & \multicolumn{4}{c}{\bf English} & \multicolumn{5}{c}{\bf Chinese} \\
\cmidrule(lr){3-6} \cmidrule(lr){7-11}
& &{\bf PPL \(\downarrow\)} & {\bf ARC-E} & {\bf Hellaswag} & {\bf Acc. Avg.} &{\bf PPL \(\downarrow\)} & {\bf GK.} & {\bf NLU} & {\bf Reasoning} & {\bf Acc. Avg.} \\ 
\midrule
Original Data & 0M &11.3 & 66.9 & 50.7 & 58.8 &41.2 & 29.8 & 52.8 & 41.6 & 41.4 \\
+Monolingual & 2,000M &\textbf{11.2} & \textbf{68.5} & 50.0 & 59.3 &\textbf{29.0} & 31.0 & 54.8 & 43.2 & 43.0 \\
\midrule
+SynCS &&&&&&&& \\
\multicolumn{1}{r}{En-Token-Repl.} & 100M &11.3 & 67.9 & \textbf{50.8} & 59.3 &38.5 & 30.8 & 55.4 & 43.0 & 43.1 (+0.0) \\
\multicolumn{1}{r}{En-Token-Repl.} & 2,000M &11.4 & 68.1 & 50.2 & 59.1 &35.0 & 31.5 & 55.4 & 47.6 & 44.9 (+1.9) \\
\multicolumn{1}{r}{Equal} & 2,000M &11.8 & 68.2 & 49.9 & 59.1 &40.5 & 30.6 & 56.1 & 46.9 & 44.5 (+1.5) \\
\multicolumn{1}{r}{Extreme} & 2,000M &11.6 & 67.9 &50.3 &59.1 &36.4 & 30.7 & 56.2 & 47.4 & 44.7 (+1.7) \\
\multicolumn{1}{r}{En-Repl. Equal} & 2,000M &11.4 & 68.4 & 50.3 &\textbf{59.4} &34.1 & \textbf{31.7} & \textbf{57.4} & \textbf{47.9} & \textbf{45.7 (+2.7)} \\
\bottomrule
\end{tabular}}
\caption{Evaluation results of different code-switching mixing strategies. ``En-Token-Repl.'' represents Token-Level-Replacement code-switching in English, which performs the best in the scaling experiments.}
\label{tab:mxing}
\end{table*}
\subsection{Bring All Together}
To investigate potential mutual promotion effects between different code-switching types and identify the optimal mixing strategy, we merge all types in both English and Chinese. 
For simplicity, code-switching of type X in language L is denoted as ``L-X''. ``En-T-Repl.'' refers to conduct Token-Level-Replacement code-switching in the English data.
Under the 500M and 2,000M budgets explored in the scaling experiments, we implement the following heuristic mixing strategies:
\begin{itemize}
    \item Equal: In each language, four types of code-switching are evenly mixed.
    \item Extreme: In each language, the most powerful type of code-switching is used at its optimal scale (En-Token-Repl. at 2,000M, and Zh-T-Annt. at 200M).
    \item En-Repl. Equal: En-Token-Repl. and En-Sent-Repl. are evenly mixed with each at the 1,000M scale, derived from their superior performance in the scaling experiments.
\end{itemize}
We expand our evaluation to three dimensions: General Knowledge (GK.), Natural Language Understanding (NLU), and Reasoning with each containing 4 benchmarks~\citep{kydlicek2024finetasksmultilingualtasks}. Details are illustrated in section~\ref{eval}.
Table~\ref{tab:mxing} presents the results.

\paragraph{SynCS Achieves 20x the Efficiency of Monolingual Data}
SynCS-Equal leads to a significant improvement (+3.16) and substantially outperforms adding an equal amount of monolingual data with natural code-switching (+1.52). 
Using the best En-Token-Repl. type at the 100M scale even demonstrates comparable performance to adding 20x monolingual data (43.1 vs 43.0).

\begin{table*}[t]
\centering
\scriptsize
\scalebox{1}{
\begin{tabular}{lccccccccc}
\toprule
\multirow{2}{*}{\bf Data} & \multirow{2}{*}{\bf \# New Tokens} & \multicolumn{4}{c}{\bf English} & \multicolumn{4}{c}{\bf Chinese} \\
\cmidrule(lr){3-6} \cmidrule(lr){7-10}
&& {\bf PPL \(\downarrow\)} &{\bf Hellaswag} &\bf ARC-E &\bf Acc. Avg. &\bf PPL \(\downarrow\) &\bf Hellaswag &\bf ARC-E &\bf Acc. Avg. \\ 
\midrule
Original Data &0M &\textbf{13.6} &\textbf{48.4} &\textbf{67.8} &\textbf{58.1} &60.0 &33.9 &49.1 &41.5 \\
+Monolingual &3,000M &13.7 &48.2 &66.4 &57.3 &\textbf{50.1} &34.6 &52.2 &43.4 \\
+SynCS* &150M &13.8 &48.4 &66.5 &57.4 &58.6 &35.1 &52.5 &43.8 \\
+SynCS* &3,000M &14.2 &46.8 &65.3 &56.1 &56.1 &\textbf{37.2} &\textbf{56.3} &\textbf{46.7} \\
\midrule
\multirow{2}{*}{\bf Data} & \multirow{2}{*}{\bf \# New Tokens} & \multicolumn{4}{c}{\bf Romanian} & \multicolumn{4}{c}{\bf Bengali} \\
\cmidrule(lr){3-6} \cmidrule(lr){7-10}
&& {\bf PPL \(\downarrow\)} &{\bf Hellaswag} &\bf ARC-E &\bf Acc. Avg. &\bf PPL \(\downarrow\) &\bf Hellaswag &\bf ARC-E &\bf Acc. Avg. \\
\midrule
Original Data &0M &9.8 &30.9 &33.9 &32.4 &9.7 &27.0 &28.9 &28.0 \\
+Monolingual &3,000M &\textbf{8.6} &32.0 &35.6 &33.8 &\textbf{7.9} &27.6 &31.5 &29.6 \\
+SynCS* &150M &9.2 &30.9 &37.1 &34.0 &8.6 &27.8 &30.1 &29.0 \\
+SynCS* &3,000M &8.7 &\textbf{32.5} &\textbf{40.7} &\textbf{36.6} &8.2 &\textbf{28.1} &\textbf{32.6} &\textbf{30.3} \\
\bottomrule
\end{tabular}}
\caption{Evaluation results in the multilingual setting.}
\label{tab:multilingual}
\end{table*}

\paragraph{Mixing SynCS in Both Languages Brings No Improvement}
Results show that SynCS-Equal and SynCS-Extreme demonstrate a slight decrease compared to En-Token-Repl., indicating that mixing SynCS in both languages does not yield significant mutual promotion effects.

\paragraph{The Most Two Powerful Types Promote Each Other}
En-Repl. Equal showcases substantial improvements over other mixing strategies. Its performance outperforms each of its composition types at the same scale, indicating the potential mutual promotion effects.
We use this strategy as our final method in the following experiments, denoted as SynCS*.
Figure~\ref{figure:mexa} shows the MEXA alignment scores. SynCS* significantly enhances MEXA alignment across all layers, particularly in shallow and deep layers, whereas monolingual data exhibits a slower, natural alignment process.

\section{Extend to Multilingual and DownStream Cross-Lingual Tasks}
\subsection{SynCS Generalizes to Other Languages}
To assess SynCS's effectiveness in multilingual settings, we select Chinese, Romanian, and Bengali as representatives of high, medium, and low-resource languages. Details of the synthesis setup are in section~\ref{syncs_appendix}. The pre-training setup follows section~\ref{pretrain}, except that the tokenizer is changed to DeepSeek-V3~\citep{liu2024deepseek} for improved tokenization of Romanian and Bengali. Due to the lack of benchmarks for Bengali and Romanian, we evaluate only on perplexity, Hellaswag, and ARC-Easy.

We first choose the same sentences at the 2,000M setting in our scaling experiments and evenly allocate them to these languages.
Notably, the total number of new low-resource language tokens becomes 3,000M beacause of the different tokenization for languages.
Table~\ref{tab:multilingual} presents that SynCS significantly outperforms the addition of an equivalent amount of monolingual documents across all three languages.
Meanwhile, the 20x efficiency ratio still holds true on Romanian.
For Bengali, SynCS presents comparable performance to its 20x monolingual data.
This demonstrates the robust language generalization capabilities of SynCS. 

\subsection{Investigation for DownStream Cross-Lingual Tasks}
To assess whether pre-training on SynCS data enhances cross-lingual transfer capabilities in downstream tasks, we conducted experiments on translation and Zero-shot Cross-lingual Transfer (ZS-CLT).

For the translation task, the models are further fine-tuned using OPUS En->Zh data and evaluated on the Flores En->Zh translation task. We measure performance using sacreBLEU~\citep{post-2018-call} and COMET~\citep{rei-etal-2022-comet}~\footnote{We employed the \textrm{wmt22-comet-da} version.}. For the ZS-CLT task, we chose XNLI~\citep{conneau-etal-2018-xnli} as the training dataset, where pre-trained models were exclusively fine-tuned on the English train set of XNLI and subsequently evaluated on both English and Chinese test sets. 

As illustrated in Table~\ref{tab:sft}, our SynCS model delivers considerable improvements in translation and ZS-CLT tasks over models trained on monolingual data, indicating that pre-training with SynCS data augments the base model's cross-lingual transfer capabilities.

\begin{table}[t]
\centering
\footnotesize
\scalebox{1}{
\begin{tabular}{lcccc}
\toprule
\multirow{2}{*}{\bf Model} & \multicolumn{2}{c}{\bf Flores (En->Zh)} & \multicolumn{2}{c}{\bf ZS-CLT} \\
\cmidrule(lr){2-3} \cmidrule(lr){4-5}
& {\bf BLEU} & {\bf COMET} & {\bf En} & {\bf Zh} \\ 
\midrule
Mono & 18.37 & 73.19 & \textbf{80.46} & 67.31 \\
SynCS & \textbf{21.81} & \textbf{76.87} & 80.38 & \textbf{70.78} \\
\bottomrule
\end{tabular}}
\caption{Evaluation results on translation task and ZS-CLT. "Mono" and "SynCS" refer to the models finetuned from "+Monolingual 2000M" and "En-Repl. Equal 2000M" in Table~\ref{tab:mxing} respectively.}
\label{tab:sft}
\end{table}

%% file: realated_work.tex
\section{Related Work}
\subsection{Cross-Lingual Transfer}
Due to the imbalance of languages in the pre-training corpora, LLMs' multilingual abilities still show significant disparities~\citep{bai2023qwen,dubey2024llama}. Since addressing this language data imbalance is challenging~\citep{ranta2021linguistic}, many efforts have been made to explore cross-lingual transfer in LLMs, which aim to transfer knowledge or reasoning capabilities from high-resource languages to low-resource languages. In the post-training stage, 
~\citet{she-etal-2024-mapo} utilize response consistency between low- and high-resource languages to optimize and enhance LLMs' multilingual reasoning using DPO or PPO. ~\citet{zhou2024moe} propose to prevent high-resource languages' catastrophic forgetting during continual pre-training for better low-resource language adaptation. In the pre-training stage, ~\citet{dufter-schutze-2020-identifying} identify shared parameters, sub-words, and position embeddings as keys to transformer's multilingualism. ~\citet{li-etal-2024-prealign} argue that aligning multilingual representations before large-scale pre-training, followed by input-only code-switching, enhances multilingual capabilities.

\subsection{Code-Switching}
Code-switching, or language alternation, is a linguistic phenomenon where multilingual speakers use multiple languages within a conversation~\citep{poplack1978syntactic}. While LLMs exhibit strong multilingual capabilities, they struggle with code-switching tasks. \citet{yoosafe} show that code-switching attack prompts increase success rates. Code-switching aids multilingual alignment, as demonstrated by \citet{li-etal-2024-prealign}, who use input-only code-switching during pre-training. \citet{yoo2024code} introduce CSCL, a curriculum learning method using synthetic code-switching data to enhance multilingual alignment. \citet{yoo2024code} is the most similar work to us. However, we focus on the pre-training stage, analyzing how natural code-switching enhances LLMs' multilingual capabilities and proposing a more flexible and less expensive code-switching synthesis approach.

%% file: conclusion.tex
\section{Conclusion}
This study explores the impact of code-switching on cross-lingual transfer during pre-training. We find that natural code-switching significantly enhances the multilingual capabilities of LLMs under extreme language imbalance. To address the scarcity of natural code-switching, we introduce a synthetic framework requiring only a small set of high-quality parallel sentences. Through extensive experiments and analysis, we demonstrate that this framework outperform those trained on equivalent monolingual data, improving performance across languages of varying resources.

%% file: limitation.tex
\section{Limitations}
Due to the resource limit, our models fall under a 1.5B small language model trained on 60B tokens, which lacks generation abilities. Whether the findings in the paper hold on larger settings remains to be explored. Table~\ref{tab:multilingual} demonstrates that the improvement achieved on the low-resource language is not substantial because of the low-quality of the pre-training and synthetic code-switching data. How to generate high-quality code-switching data for these languages is a problem. Additionally, models trained with SynCS demonstrates worse performance on the Wiki-ppl compared to monolingual data, which may be handled by continue training on monolingual data or using the input-only code-switching~\citep{li-etal-2024-prealign}.
We leave these limitations for further work.

\section{Acknowledgement}
We would like to thank the anonymous reviewers for their insightful comments. Shujian Huang is the corresponding author. This work is supported by National Science Foundation of China (No. 62376116, 62176120), research project of Nanjing University-China Mobile Joint Institute, the Fundamental Research Funds for the Central Universities (No. 2024300507).

%% file: appendix.tex
\appendix
\label{appendix}

\section{Code-Switching Data Detecting}
\subsection{Detecting Details}
\label{detecting}
We first apply a character-based filtering to obtain documents that contain English and Chinese.
Then we use fasttext~\citep{joulin2017bag} to classify each sentence as monolingual or bilingual, corresponding to sentence-level and token-level code-switching, respectively. 
We prompt Qwen2.5-72B-Instruct~\citep{yang2024qwen2} to filter out the unrelated code-switching sentences.
Each segment is then categorized as either Annt. or Repl..

For sentence level, classifying into Annt. and Repl. is indeed detecting the translation pairs.
We employ LABSE~\citep{feng-etal-2022-language} cross-lingual encoder to find semantic-align sentence pairs in two languages, following~\citet{briakou-etal-2023-searching}.

For token level, we use an LLM-based detection strategy to categorize. We prompt Qwen2.5-72B-Instruct with the instructions as following and ask for classification.

\begin{tcolorbox}[breakable, colback=gray!5!white, colframe=gray!75!black, title=Prompts for Annotation and Replacement classification]\label{detect_qwen}
Code-switching can be classified more finely according to different characteristics and uses. Here are some common types:\\

1. Annotation: In this case, another language is used to explain or define a noun before or after it. For example: During the festival, we watched a dragon dance (舞龙). In this sentence, the word "舞龙" serves as an annotation for "dragon dance".

2. Replacement: A specific word is replaced by a foreign word. For example: During the festival, we watched a 舞龙. In this sentence, the word "舞龙" replaces the English word "dragon dance".\\

Given an English sentence containing Chinese code-switching, please classify the sentence according to the above two types.

Examples:

[English Sentence]: During the festival, we watched a dragon dance (舞龙), which is a traditional Chinese performance.

[Answer]: "舞龙" appears after "dragon dance", which explains this English word in Chinese and is its annotation. Formatting result: \textbackslash \textbackslash box(1)\\

[English Sentence]: We enjoyed some delicious food at a nearby 茶馆.

[Answer]: The word "茶馆" is directly used as part of the sentence. It can be assumed that the original word is "teahouse", but it is directly replaced by "茶馆". Formatting result: \textbackslash \textbackslash box(2)\\

The following is your task. You can do a brief analysis, but please be sure to output it in the format of the example at the end.

[English Sentence]:

[Answer]:
\end{tcolorbox}

\section{Examples for Various Natural Code-Switching Segments}
\begin{tcolorbox}[colback=gray!5!white, colframe=gray!75!black, title=English-Side Code-Switching]
\label{En-side-natural}
Unrelated:\\
1. ◇ お客様、こちらのブラウスですと、いまお召しのスーツにもよく合いますが。\\
2. there are also the phrases いつ頃 (about when? \\
3. 2 Polypodiaceae Tac ke 家 Me.

\end{tcolorbox}

\begin{tcolorbox}[breakable, colback=gray!5!white, colframe=gray!75!black, title=Chinese-Side Code-Switching]
\label{zh-side-natural}
Unrelated:\\
1. zxx520llc发表于: 2个月前\#9\\
2. X\$Gx17{0 水利图书 F'} q A\\t\^8t2G [Garbled] \\

T-Annt.:\\
1.比如\uline{盐酸}(\textcolor{blue}{HCL})、硝酸。[Explanation in English: For example, hydrochloric acid (HCL) and nitric acid.] \\

T-Repl.:\\
1. \textcolor{blue}{Microsoft} 商店很可能误解了你尝试下载或安装的应用程序。[Explanation in English: It's possible that the Microsoft Store misunderstood the app you were trying to download or install.]\\

S-Annt.:\\
1. 任何人都不太可能真正了解它的全部。\textcolor{blue}{These are the basic materials that go into a pencil, graphite, cedar, metal,and rubber。}\uline{这些就是构成铅笔的基本材料，\\石墨、雪松、金属、橡胶。} \\

S-Repl.:\\
1. 我只想引述GPT-4官方新闻的一句话：\textcolor{blue}{As a result, our GPT-4 training run was (for us at least!) unprecedentedly stable.} [Explanation in English: I just want to quote a sentence from the official GPT-4 news: As a result, our GPT-4 training run was (for us at least!) unprecedentedly stable.]\\

\end{tcolorbox}

\section{Code-Switching Data Synthesis}
\label{syncs_appendix}
\paragraph{Synthesis Model Training Details}
We utilize 4 A100 GPUs and conduct multilingual and multi-task supervised fine-tuning on Qwen2.5-3B-Instruct. The model is fine-tuned for 3 epochs, using a context length of 2048 tokens, a warmup ratio of 0.1, and a peak of learning rate at 5e-5 with cosine decaying to 0. We utilize bf16 mixed precision and flash attention~\citep{dao2023flashattention2} to speed up the training process. We assign the temperature as 0 when generating code-switching data and translating sentences (i.e. greedy decoding). vLLM~\citep{kwon2023efficient} is used to accelerate the generation.

The source data for generating code-switching supervised fine-tuning data includes X-ALMA~\citep{xu2024x} and flores200~\citep{costa2022no}.
While TowerInstruct doesn't support Bengali, we use NLLB~\citep{costa2022no} as the translator.
As the data of ~\citet{xu2024x} doesn't contain Bengali, we directly use the dev and devtest set of the flores200~\citep{costa2022no} dataset.
Table~\ref{tab:orcaledata} shows the number of parallel sentences in each language when generating the SFT data. We use the same data for the Annotation and Replacement types in both languages, resulting in a total of 62000 multilingual and multi-task SFT data. We directly reuse the prompts above except only the source language sentence is given.

\subsection{Synthesis Prompts}
When generating the token-level code-switching SFT data using GPT4o-mini, we follow and slightly modify the prompt of ~\citet{yoo2024code} for better instruction-following.

\begin{table}[t]
\centering
\scalebox{1}{
\begin{tabular}{lc}
\toprule
{\bf Language Pairs} &{\# of Parallel Sentences} \\
\midrule
English-Chinese &6906 \\
English-Romanian &4987 \\
English-Bengali &3604 \\
Total &15500 \\
\bottomrule
\end{tabular}}
\caption{Number of parallel sentences used for generating token-level code-switching SFT data.}
\label{tab:orcaledata}
\end{table}

\begin{tcolorbox}[colback=gray!5!white, colframe=gray!75!black, title=Prompts of Code-Switching Generation]
\label{prompts_syncs}
Annotation (Target-Side as example):

Given a pair of \{\textit{Source Language}\}-English parallel sentence, generate an English-annotated \{\textit{Source Language}\} sentence. Annotation is the use of words from another language to explain certain words in a sentence. 

[\{\textit{Source Language}\} Sentence]: 
\\

Replacement:

Given a pair of \{\textit{Source Language}\}-English sentence, generate a \{\textit{Source Language}\} and English code-switching sentence. Code-switching is the use of more than one linguistic variety in a manner consistent with the syntax and phonology of each variety.

[\{\textit{Source Language}\} Sentence]: 
\end{tcolorbox}

\section{Experiment Settings}
\label{setup}
\paragraph{Pre-Training Recipes}
\label{pretrain}
We sample 60B English tokens from FineWeb-Edu and 600M Chinese tokens from Chinese-FineWeb-Edu-v2 to simulate the language-imbalance (100:1) pre-training. A 1.5B Qwen2.5 model~\citep{yang2024qwen2} is trained on this sampled data to explore the cross-lingual transfer during pre-training. All models are trained for 30,000 steps with a batch size of 2M tokens. We group training documents with the length of 2048 and pre-training with global batch size of 1024. The learning rate performs cosine decay from 2e-4 to 5e-6 with 1\% warmup. Experiments are conducted on the Megatron-LM~\citep{shoeybi2019megatron} framework. We use flash-attn~\citep{dao2023flashattention2} to accelerate training. Each experiment is trained on 128 A100s for 9 hours.

\paragraph{Evaluation Recipes}
\label{eval}
We use the perplexity on Wikipedia~\citep{wikidump} and the finetasks~\citep{kydlicek2024finetasksmultilingualtasks} to evaluate our models.
In finetasks, we choose the 12 tasks belonging to 3 dimensions: 
\begin{itemize}
    \item General Knowledge: AGI-Eval~\citep{zhong-etal-2024-agieval}, C-EVAL~\citep{huang2023ceval}, CMMLU~\citep{li-etal-2024-cmmlu}, M3Exams~\citep{zhang2023m3exam}.
    \item Natural Langauge Understanding: M-Hellaswag~\citep{lai-etal-2023-okapi}, Ocnli~\citep{ocnli}, X-winigrad~\citep{muennighoff-etal-2023-crosslingual}, Xstory-cloze~\citep{mostafazadeh2017lsdsem}.
    \item Reasoning: Xcodah~\citep{chen-etal-2019-codah}, XCOPA~\citep{ponti-etal-2020-xcopa}, XCSPA~\citep{lin-etal-2021-common}, ARC-Easy~\citep{clark2018think}.
\end{itemize}

The multilingual translated version of Hellaswag~\citep{lai-etal-2023-okapi} is used.
Since there is no multilingual version of ARC-Easy, we translate the original English version to Chinese, Romanian, and Bengali using GPT-4o-mini, following~\citet{lai-etal-2023-okapi}.
We also present MEXA~\citep{kargaran2024mexa} scores, which assess alignment between English and non-English languages using parallel sentences, flores200~\citep{costa2022no}, to evaluate language transfer.
When we explore the natural code-switching and scaling up the synthetic code-switching, since the differences on these benchmarks are insignificant at a small scale, only perplexity, Hellaswag, and ARC-Easy are reported.
Besides, in our multilingual settings, there are lack of evaluation benchmarks for Bengali and Romanian.
We also only report these three results.

\section{T-SNE Visualization}
\begin{figure*}[htb]
\centering
\includegraphics[width=0.85\textwidth]{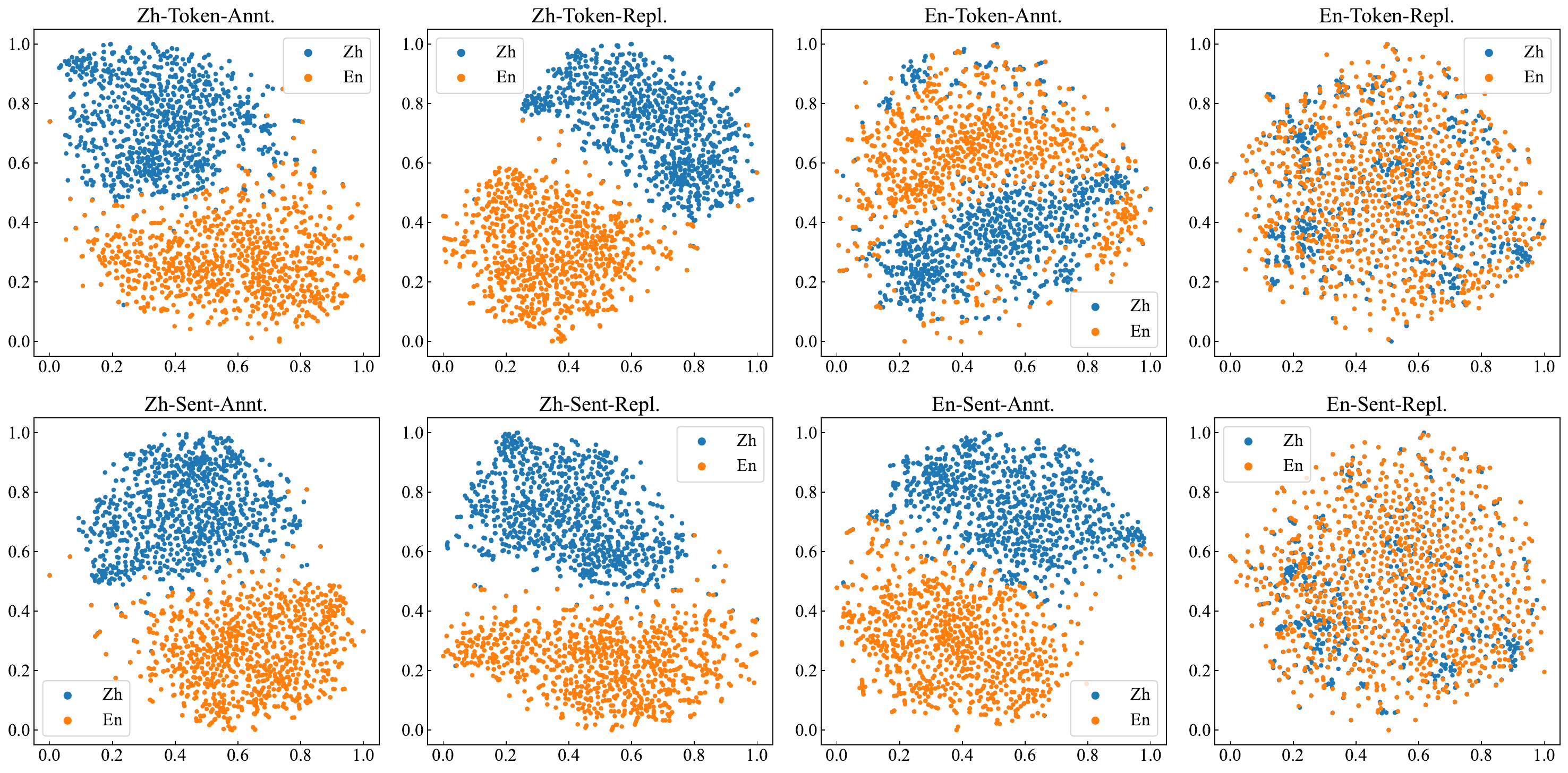}
\caption{T-SNE visualization of parallel sentences' middle layer hidden states for models trained on Chinese-side and English-side SynCS.}
\label{figure:all_tsne}
\end{figure*}

Figure~\ref{figure:all_tsne} demonstrates the T-SNE visualization of parallel sentences' middle layer hidden states for models trained on Chinese and English-side SynCS respectively.
Only En-Token-Repl. and En-S-Repl. showcase obvious differences for mixing the representation space in two languages.

\subsection{Detailed Evaluations}
Table~\ref{tab:gk}, \ref{tab:nlu}, and \ref{tab:reasoning} presents the detailed evaluations on each Chinese benchmarks mentioned at Table~\ref{tab:mxing}.

\begin{table*}[ht]
\centering
\footnotesize
\scalebox{1}{
\begin{tabular}{lcccccccccc}
\toprule
\textbf{Data} &\bf \# New Tokens &\bf AGI-Eval &\bf CEVAL &\bf CMMLU &\bf M3Exams &\bf Avg. \\
\midrule
Original Data & 0M &28.8 &28.3 &30.1 &31.9 &29.8\\
+Monolingual & 2,000M &29.5 &31.0 &31.6 &32.0 &31.0 \\
\midrule
+SynCS &&&&&&&& \\
\multicolumn{1}{r}{En-Token-Repl.} & 100M & 30.5     & 30.2   & 30.8   & 31.9     & 30.8 \\
\multicolumn{1}{r}{En-Token-Repl.} & 2,000M & 30.7     & 31.3   & 31.8   & 32.3     & 31.5 \\
\multicolumn{1}{r}{Equal} & 2,000M & 29.7     & 29.5   & 30.6   & 32.8 &30.6 \\
\multicolumn{1}{r}{Extreme} & 2,000M & 29.2     & 30.9   & 31.1   & 31.4 &30.7 \\
\multicolumn{1}{r}{En-Repl. Equal} & 2,000M & 30.5     & 29.9   & 31.5   & 35.1     &31.7 \\
\bottomrule
\end{tabular}}
\caption{Chinese evaluation results on the General Knowledge (GK.) evaluation set.}
\label{tab:gk}
\end{table*}

\begin{table*}[ht]
\centering
\footnotesize
\scalebox{1}{
\begin{tabular}{lcccccccccc}
\toprule
\textbf{Data} &\bf \# New Tokens &\bf AGI-Eval &\bf CEVAL &\bf CMMLU &\bf M3Exams &\bf Avg. \\
\midrule
Original Data & 0M &33.8 & 54.3 & 65.5 & 57.8 & 52.8 \\
+Monolingual & 2,000M &35.3 & 56.8 & 66.9 & 60.3 & 54.8 \\
\midrule
+SynCS &&&&&&&& \\
\multicolumn{1}{r}{En-Token-Repl.} & 100M &35.8 & 59.9 & 67.7 & 58.3 & 55.4 \\
\multicolumn{1}{r}{En-Token-Repl.} & 2,000M &39.7 & 55.2 & 68.9 & 57.7 & 55.4 \\
\multicolumn{1}{r}{Equal} & 2,000M &38.5 & 60.3 & 66.7 & 59.1 & 56.1 \\
\multicolumn{1}{r}{Extreme} & 2,000M &39.2 & 58.3 & 66.9 & 60.3 & 56.2 \\
\multicolumn{1}{r}{En-Repl. Equal} & 2,000M &40.1 & 62.4 & 66.9 & 60.2 & 57.4 \\
\bottomrule
\end{tabular}}
\caption{Chinese evaluation results on the Natural Language Understanding (NLU) evaluation set.}
\label{tab:nlu}
\end{table*}

\begin{table*}[ht]
\centering
\footnotesize
\scalebox{1}{
\begin{tabular}{lcccccccccc}
\toprule
\textbf{Data} &\bf \# New Tokens &\bf XCodah &\bf XCOPA &\bf XCSQA &\bf ARC-Easy &\bf Avg. \\
\midrule
Original Data & 0M &33.0 & 57.4 & 35.9 & 39.9 & 41.6 \\
+Monolingual & 2,000M &33.0 & 58.6 & 36.3 & 45.0 & 43.2 \\
\midrule
+SynCS &&&&&&&& \\
\multicolumn{1}{r}{En-Token-Repl.} & 100M &33.7 & 56.6 & 35.4 & 46.5 & 43.0 \\
\multicolumn{1}{r}{En-Token-Repl.} & 2,000M &35.7 & 61.8 & 39.0 & 54.1 & 47.6 \\
\multicolumn{1}{r}{Equal} & 2,000M &32.7 & 62.4 & 38.3 & 54.0 & 46.9 \\
\multicolumn{1}{r}{Extreme} & 2,000M &34.3 & 60.0 & 40.0 & 55.2 & 47.4 \\
\multicolumn{1}{r}{En-Repl. Equal} & 2,000M &33.3 & 61.4 & 40.0 & 56.8 & 47.9 \\
\bottomrule
\end{tabular}}
\caption{Chinese evaluation results on the Reasoning evaluation set.}
\label{tab:reasoning}
\end{table*}

%% file: acl_latex.bbl
\begin{thebibliography}{51}
\providecommand{\natexlab}[1]{#1}

\bibitem[{Achiam et~al.(2023)Achiam, Adler, Agarwal, Ahmad, Akkaya, Aleman, Almeida, Altenschmidt, Altman, Anadkat et~al.}]{achiam2023gpt}
Josh Achiam, Steven Adler, Sandhini Agarwal, Lama Ahmad, Ilge Akkaya, Florencia~Leoni Aleman, Diogo Almeida, Janko Altenschmidt, Sam Altman, Shyamal Anadkat, et~al. 2023.
\newblock Gpt-4 technical report.
\newblock \emph{arXiv preprint arXiv:2303.08774}.

\bibitem[{Bai et~al.(2023)Bai, Bai, Chu, Cui, Dang, Deng, Fan, Ge, Han, Huang et~al.}]{bai2023qwen}
Jinze Bai, Shuai Bai, Yunfei Chu, Zeyu Cui, Kai Dang, Xiaodong Deng, Yang Fan, Wenbin Ge, Yu~Han, Fei Huang, et~al. 2023.
\newblock Qwen technical report.
\newblock \emph{arXiv preprint arXiv:2309.16609}.

\bibitem[{Briakou et~al.(2023)Briakou, Cherry, and Foster}]{briakou-etal-2023-searching}
Eleftheria Briakou, Colin Cherry, and George Foster. 2023.
\newblock \href {https://doi.org/10.18653/v1/2023.acl-long.524} {Searching for needles in a haystack: On the role of incidental bilingualism in {P}a{LM}`s translation capability}.
\newblock In \emph{Proceedings of the 61st Annual Meeting of the Association for Computational Linguistics (Volume 1: Long Papers)}, pages 9432--9452, Toronto, Canada. Association for Computational Linguistics.

\bibitem[{Chen et~al.(2019)Chen, D{'}Arcy, Liu, Fernandez, and Downey}]{chen-etal-2019-codah}
Michael Chen, Mike D{'}Arcy, Alisa Liu, Jared Fernandez, and Doug Downey. 2019.
\newblock \href {https://doi.org/10.18653/v1/W19-2008} {{CODAH}: An adversarially-authored question answering dataset for common sense}.
\newblock In \emph{Proceedings of the 3rd Workshop on Evaluating Vector Space Representations for {NLP}}, pages 63--69, Minneapolis, USA. Association for Computational Linguistics.

\bibitem[{Clark et~al.(2018)Clark, Cowhey, Etzioni, Khot, Sabharwal, Schoenick, and Tafjord}]{clark2018think}
Peter Clark, Isaac Cowhey, Oren Etzioni, Tushar Khot, Ashish Sabharwal, Carissa Schoenick, and Oyvind Tafjord. 2018.
\newblock Think you have solved question answering? try arc, the ai2 reasoning challenge.
\newblock \emph{arXiv preprint arXiv:1803.05457}.

\bibitem[{Colombo et~al.(2024)Colombo, Alves, Pombal, Guerreiro, Martins, Alves, Farajian, Peters, Rei, Fernandes et~al.}]{colombo2024tower}
Pierre Colombo, Duarte Alves, Jos{\'e} Pombal, Nuno Guerreiro, Pedro Martins, Joao Alves, Amin Farajian, Ben Peters, Ricardo Rei, Patrick Fernandes, et~al. 2024.
\newblock Tower: An open multilingual large language model for translation-related tasks.

\bibitem[{Conneau et~al.(2018)Conneau, Rinott, Lample, Williams, Bowman, Schwenk, and Stoyanov}]{conneau-etal-2018-xnli}
Alexis Conneau, Ruty Rinott, Guillaume Lample, Adina Williams, Samuel Bowman, Holger Schwenk, and Veselin Stoyanov. 2018.
\newblock \href {https://doi.org/10.18653/v1/D18-1269} {{XNLI}: Evaluating cross-lingual sentence representations}.
\newblock In \emph{Proceedings of the 2018 Conference on Empirical Methods in Natural Language Processing}, pages 2475--2485, Brussels, Belgium. Association for Computational Linguistics.

\bibitem[{Costa-juss{\`a} et~al.(2022)Costa-juss{\`a}, Cross, {\c{C}}elebi, Elbayad, Heafield, Heffernan, Kalbassi, Lam, Licht, Maillard et~al.}]{costa2022no}
Marta~R Costa-juss{\`a}, James Cross, Onur {\c{C}}elebi, Maha Elbayad, Kenneth Heafield, Kevin Heffernan, Elahe Kalbassi, Janice Lam, Daniel Licht, Jean Maillard, et~al. 2022.
\newblock No language left behind: Scaling human-centered machine translation.
\newblock \emph{arXiv preprint arXiv:2207.04672}.

\bibitem[{Dao(2024)}]{dao2023flashattention2}
Tri Dao. 2024.
\newblock Flash{A}ttention-2: Faster attention with better parallelism and work partitioning.
\newblock In \emph{International Conference on Learning Representations (ICLR)}.

\bibitem[{Dubey et~al.(2024)Dubey, Jauhri, Pandey, Kadian, Al-Dahle, Letman, Mathur, Schelten, Yang, Fan et~al.}]{dubey2024llama}
Abhimanyu Dubey, Abhinav Jauhri, Abhinav Pandey, Abhishek Kadian, Ahmad Al-Dahle, Aiesha Letman, Akhil Mathur, Alan Schelten, Amy Yang, Angela Fan, et~al. 2024.
\newblock The llama 3 herd of models.
\newblock \emph{arXiv preprint arXiv:2407.21783}.

\bibitem[{Dufter and Sch{\"u}tze(2020)}]{dufter-schutze-2020-identifying}
Philipp Dufter and Hinrich Sch{\"u}tze. 2020.
\newblock \href {https://doi.org/10.18653/v1/2020.emnlp-main.358} {Identifying elements essential for {BERT}`s multilinguality}.
\newblock In \emph{Proceedings of the 2020 Conference on Empirical Methods in Natural Language Processing (EMNLP)}, pages 4423--4437, Online. Association for Computational Linguistics.

\bibitem[{Feng et~al.(2022)Feng, Yang, Cer, Arivazhagan, and Wang}]{feng-etal-2022-language}
Fangxiaoyu Feng, Yinfei Yang, Daniel Cer, Naveen Arivazhagan, and Wei Wang. 2022.
\newblock \href {https://doi.org/10.18653/v1/2022.acl-long.62} {Language-agnostic {BERT} sentence embedding}.
\newblock In \emph{Proceedings of the 60th Annual Meeting of the Association for Computational Linguistics (Volume 1: Long Papers)}, pages 878--891, Dublin, Ireland. Association for Computational Linguistics.

\bibitem[{Foundation()}]{wikidump}
Wikimedia Foundation.
\newblock \href {https://dumps.wikimedia.org} {Wikimedia downloads}.

\bibitem[{Hu et~al.(2020)Hu, Richardson, Xu, Li, Kuebler, and Moss}]{ocnli}
Hai Hu, Kyle Richardson, Liang Xu, Lu~Li, Sandra Kuebler, and Larry Moss. 2020.
\newblock \href {https://arxiv.org/abs/2010.05444} {Ocnli: Original chinese natural language inference}.
\newblock In \emph{Findings of EMNLP}.

\bibitem[{Huang et~al.(2023)Huang, Bai, Zhu, Zhang, Zhang, Su, Liu, Lv, Zhang, Lei, Fu, Sun, and He}]{huang2023ceval}
Yuzhen Huang, Yuzhuo Bai, Zhihao Zhu, Junlei Zhang, Jinghan Zhang, Tangjun Su, Junteng Liu, Chuancheng Lv, Yikai Zhang, Jiayi Lei, Yao Fu, Maosong Sun, and Junxian He. 2023.
\newblock C-eval: A multi-level multi-discipline chinese evaluation suite for foundation models.
\newblock In \emph{Advances in Neural Information Processing Systems}.

\bibitem[{Joulin et~al.(2017)Joulin, Grave, Bojanowski, and Mikolov}]{joulin2017bag}
Armand Joulin, Edouard Grave, Piotr Bojanowski, and Tomas Mikolov. 2017.
\newblock Bag of tricks for efficient text classification.
\newblock In \emph{Proceedings of the 15th Conference of the European Chapter of the Association for Computational Linguistics: Volume 2, Short Papers}, pages 427--431. Association for Computational Linguistics.

\bibitem[{Kargaran et~al.(2024)Kargaran, Modarressi, Nikeghbal, Diesner, Yvon, and Sch{\"u}tze}]{kargaran2024mexa}
Amir~Hossein Kargaran, Ali Modarressi, Nafiseh Nikeghbal, Jana Diesner, Fran{\c{c}}ois Yvon, and Hinrich Sch{\"u}tze. 2024.
\newblock Mexa: Multilingual evaluation of english-centric llms via cross-lingual alignment.
\newblock \emph{arXiv preprint arXiv:2410.05873}.

\bibitem[{Kwon et~al.(2023)Kwon, Li, Zhuang, Sheng, Zheng, Yu, Gonzalez, Zhang, and Stoica}]{kwon2023efficient}
Woosuk Kwon, Zhuohan Li, Siyuan Zhuang, Ying Sheng, Lianmin Zheng, Cody~Hao Yu, Joseph~E. Gonzalez, Hao Zhang, and Ion Stoica. 2023.
\newblock Efficient memory management for large language model serving with pagedattention.
\newblock In \emph{Proceedings of the ACM SIGOPS 29th Symposium on Operating Systems Principles}.

\bibitem[{Kydlíček et~al.()Kydlíček, Penedo, Fourier, Habib, and Wolf}]{kydlicek2024finetasksmultilingualtasks}
Hynek Kydlíček, Guilherme Penedo, Clémentine Fourier, Nathan Habib, and Thomas Wolf.
\newblock \href {https://huggingface.co/spaces/HuggingFaceFW/blogpost-fine-tasks} {Finetasks: Finding signal in a haystack of 200+ multilingual tasks}.

\bibitem[{Lai et~al.(2023)Lai, Nguyen, Ngo, Nguyen, Dernoncourt, Rossi, and Nguyen}]{lai-etal-2023-okapi}
Viet Lai, Chien Nguyen, Nghia Ngo, Thuat Nguyen, Franck Dernoncourt, Ryan Rossi, and Thien Nguyen. 2023.
\newblock \href {https://doi.org/10.18653/v1/2023.emnlp-demo.28} {Okapi: Instruction-tuned large language models in multiple languages with reinforcement learning from human feedback}.
\newblock In \emph{Proceedings of the 2023 Conference on Empirical Methods in Natural Language Processing: System Demonstrations}, pages 318--327, Singapore. Association for Computational Linguistics.

\bibitem[{Li et~al.(2024{\natexlab{a}})Li, Zhang, Koto, Yang, Zhao, Gong, Duan, and Baldwin}]{li-etal-2024-cmmlu}
Haonan Li, Yixuan Zhang, Fajri Koto, Yifei Yang, Hai Zhao, Yeyun Gong, Nan Duan, and Timothy Baldwin. 2024{\natexlab{a}}.
\newblock \href {https://doi.org/10.18653/v1/2024.findings-acl.671} {{CMMLU}: Measuring massive multitask language understanding in {C}hinese}.
\newblock In \emph{Findings of the Association for Computational Linguistics: ACL 2024}, pages 11260--11285, Bangkok, Thailand. Association for Computational Linguistics.

\bibitem[{Li et~al.(2024{\natexlab{b}})Li, Huang, Ching, Dai, and Chen}]{li-etal-2024-prealign}
Jiahuan Li, Shujian Huang, Aarron Ching, Xinyu Dai, and Jiajun Chen. 2024{\natexlab{b}}.
\newblock \href {https://doi.org/10.18653/v1/2024.emnlp-main.572} {{P}re{A}lign: Boosting cross-lingual transfer by early establishment of multilingual alignment}.
\newblock In \emph{Proceedings of the 2024 Conference on Empirical Methods in Natural Language Processing}, pages 10246--10257, Miami, Florida, USA. Association for Computational Linguistics.

\bibitem[{Lin et~al.(2021)Lin, Lee, Qiao, and Ren}]{lin-etal-2021-common}
Bill~Yuchen Lin, Seyeon Lee, Xiaoyang Qiao, and Xiang Ren. 2021.
\newblock \href {https://doi.org/10.18653/v1/2021.acl-long.102} {Common sense beyond {E}nglish: Evaluating and improving multilingual language models for commonsense reasoning}.
\newblock In \emph{Proceedings of the 59th Annual Meeting of the Association for Computational Linguistics and the 11th International Joint Conference on Natural Language Processing (Volume 1: Long Papers)}, pages 1274--1287, Online. Association for Computational Linguistics.

\bibitem[{Liu et~al.(2024)Liu, Feng, Xue, Wang, Wu, Lu, Zhao, Deng, Zhang, Ruan et~al.}]{liu2024deepseek}
Aixin Liu, Bei Feng, Bing Xue, Bingxuan Wang, Bochao Wu, Chengda Lu, Chenggang Zhao, Chengqi Deng, Chenyu Zhang, Chong Ruan, et~al. 2024.
\newblock Deepseek-v3 technical report.
\newblock \emph{arXiv preprint arXiv:2412.19437}.

\bibitem[{Mostafazadeh et~al.(2017)Mostafazadeh, Roth, Louis, Chambers, and Allen}]{mostafazadeh2017lsdsem}
Nasrin Mostafazadeh, Michael Roth, Annie Louis, Nathanael Chambers, and James~F Allen. 2017.
\newblock Lsdsem 2017 shared task: The story cloze test.
\newblock In \emph{2nd Workshop on Linking Models of Lexical, Sentential and Discourse-level Semantics}, pages 46--51. Association for Computational Linguistics.

\bibitem[{Muennighoff et~al.(2023)Muennighoff, Wang, Sutawika, Roberts, Biderman, Le~Scao, Bari, Shen, Yong, Schoelkopf, Tang, Radev, Aji, Almubarak, Albanie, Alyafeai, Webson, Raff, and Raffel}]{muennighoff-etal-2023-crosslingual}
Niklas Muennighoff, Thomas Wang, Lintang Sutawika, Adam Roberts, Stella Biderman, Teven Le~Scao, M~Saiful Bari, Sheng Shen, Zheng~Xin Yong, Hailey Schoelkopf, Xiangru Tang, Dragomir Radev, Alham~Fikri Aji, Khalid Almubarak, Samuel Albanie, Zaid Alyafeai, Albert Webson, Edward Raff, and Colin Raffel. 2023.
\newblock \href {https://doi.org/10.18653/v1/2023.acl-long.891} {Crosslingual generalization through multitask finetuning}.
\newblock In \emph{Proceedings of the 61st Annual Meeting of the Association for Computational Linguistics (Volume 1: Long Papers)}, pages 15991--16111, Toronto, Canada. Association for Computational Linguistics.

\bibitem[{OpenAI(2023)}]{chatgpt}
OpenAI. 2023.
\newblock \href {https://chat.openai.com} {Chatgpt (mar 23 version) [large language model]}.

\bibitem[{Penedo et~al.(2024)Penedo, Kydl\'{\i}\v{c}ek, Ben~allal, Lozhkov, Mitchell, Raffel, Von~Werra, and Wolf}]{NEURIPS2024_370df50c}
Guilherme Penedo, Hynek Kydl\'{\i}\v{c}ek, Loubna Ben~allal, Anton Lozhkov, Margaret Mitchell, Colin~A Raffel, Leandro Von~Werra, and Thomas Wolf. 2024.
\newblock \href {https://proceedings.neurips.cc/paper_files/paper/2024/file/370df50ccfdf8bde18f8f9c2d9151bda-Paper-Datasets_and_Benchmarks_Track.pdf} {The fineweb datasets: Decanting the web for the finest text data at scale}.
\newblock In \emph{Advances in Neural Information Processing Systems}, volume~37, pages 30811--30849. Curran Associates, Inc.

\bibitem[{Pires et~al.(2019)Pires, Schlinger, and Garrette}]{pires-etal-2019-multilingual}
Telmo Pires, Eva Schlinger, and Dan Garrette. 2019.
\newblock \href {https://doi.org/10.18653/v1/P19-1493} {How multilingual is multilingual {BERT}?}
\newblock In \emph{Proceedings of the 57th Annual Meeting of the Association for Computational Linguistics}, pages 4996--5001, Florence, Italy. Association for Computational Linguistics.

\bibitem[{Ponti et~al.(2020)Ponti, Glava{\v{s}}, Majewska, Liu, Vuli{\'c}, and Korhonen}]{ponti-etal-2020-xcopa}
Edoardo~Maria Ponti, Goran Glava{\v{s}}, Olga Majewska, Qianchu Liu, Ivan Vuli{\'c}, and Anna Korhonen. 2020.
\newblock \href {https://doi.org/10.18653/v1/2020.emnlp-main.185} {{XCOPA}: A multilingual dataset for causal commonsense reasoning}.
\newblock In \emph{Proceedings of the 2020 Conference on Empirical Methods in Natural Language Processing (EMNLP)}, pages 2362--2376, Online. Association for Computational Linguistics.

\bibitem[{Poplack(1978)}]{poplack1978syntactic}
Shana Poplack. 1978.
\newblock Syntactic structure and social function of code switching.
\newblock \emph{Latino Discourse and Communicative Behavior/Ablex Publishing}.

\bibitem[{Post(2018)}]{post-2018-call}
Matt Post. 2018.
\newblock \href {https://www.aclweb.org/anthology/W18-6319} {A call for clarity in reporting {BLEU} scores}.
\newblock In \emph{Proceedings of the Third Conference on Machine Translation: Research Papers}, pages 186--191, Belgium, Brussels. Association for Computational Linguistics.

\bibitem[{Pu et~al.(2023)Pu, Gao, and Wan}]{pu2023summarization}
Xiao Pu, Mingqi Gao, and Xiaojun Wan. 2023.
\newblock Summarization is (almost) dead.
\newblock \emph{arXiv preprint arXiv:2309.09558}.

\bibitem[{Ranta and Goutte(2021)}]{ranta2021linguistic}
Aarne Ranta and Cyril Goutte. 2021.
\newblock Linguistic diversity in natural language processing.
\newblock \emph{Traitement Automatique des Langues}, 62(3):7--11.

\bibitem[{Rei et~al.(2022)Rei, C.~de Souza, Alves, Zerva, Farinha, Glushkova, Lavie, Coheur, and Martins}]{rei-etal-2022-comet}
Ricardo Rei, Jos{\'e}~G. C.~de Souza, Duarte Alves, Chrysoula Zerva, Ana~C Farinha, Taisiya Glushkova, Alon Lavie, Luisa Coheur, and Andr{\'e} F.~T. Martins. 2022.
\newblock \href {https://aclanthology.org/2022.wmt-1.52} {{COMET}-22: Unbabel-{IST} 2022 submission for the metrics shared task}.
\newblock In \emph{Proceedings of the Seventh Conference on Machine Translation (WMT)}, pages 578--585, Abu Dhabi, United Arab Emirates (Hybrid). Association for Computational Linguistics.

\bibitem[{Robinson and Wingate(2023)}]{robinson2023leveraging}
Joshua Robinson and David Wingate. 2023.
\newblock \href {https://openreview.net/forum?id=yKbprarjc5B} {Leveraging large language models for multiple choice question answering}.
\newblock In \emph{The Eleventh International Conference on Learning Representations}.

\bibitem[{She et~al.(2024)She, Zou, Huang, Zhu, Liu, Geng, and Chen}]{she-etal-2024-mapo}
Shuaijie She, Wei Zou, Shujian Huang, Wenhao Zhu, Xiang Liu, Xiang Geng, and Jiajun Chen. 2024.
\newblock \href {https://doi.org/10.18653/v1/2024.acl-long.539} {{MAPO}: Advancing multilingual reasoning through multilingual-alignment-as-preference optimization}.
\newblock In \emph{Proceedings of the 62nd Annual Meeting of the Association for Computational Linguistics (Volume 1: Long Papers)}, pages 10015--10027, Bangkok, Thailand. Association for Computational Linguistics.

\bibitem[{Shoeybi et~al.(2019)Shoeybi, Patwary, Puri, LeGresley, Casper, and Catanzaro}]{shoeybi2019megatron}
Mohammad Shoeybi, Mostofa Patwary, Raul Puri, Patrick LeGresley, Jared Casper, and Bryan Catanzaro. 2019.
\newblock Megatron-lm: Training multi-billion parameter language models using model parallelism.
\newblock \emph{arXiv preprint arXiv:1909.08053}.

\bibitem[{Thara and Poornachandran(2018)}]{8554413}
S~Thara and Prabaharan Poornachandran. 2018.
\newblock \href {https://doi.org/10.1109/ICACCI.2018.8554413} {Code-mixing: A brief survey}.
\newblock In \emph{2018 International Conference on Advances in Computing, Communications and Informatics (ICACCI)}, pages 2382--2388.

\bibitem[{Touvron et~al.(2023)Touvron, Martin, Stone, Albert, Almahairi, Babaei, Bashlykov, Batra, Bhargava, Bhosale et~al.}]{touvron2023llama}
Hugo Touvron, Louis Martin, Kevin Stone, Peter Albert, Amjad Almahairi, Yasmine Babaei, Nikolay Bashlykov, Soumya Batra, Prajjwal Bhargava, Shruti Bhosale, et~al. 2023.
\newblock Llama 2: Open foundation and fine-tuned chat models.
\newblock \emph{arXiv preprint arXiv:2307.09288}.

\bibitem[{Van~der Maaten and Hinton(2008)}]{van2008visualizing}
Laurens Van~der Maaten and Geoffrey Hinton. 2008.
\newblock Visualizing data using t-sne.
\newblock \emph{Journal of machine learning research}, 9(11).

\bibitem[{Xu et~al.(2024)Xu, Murray, Koehn, Hoang, Eriguchi, and Khayrallah}]{xu2024x}
Haoran Xu, Kenton Murray, Philipp Koehn, Hieu Hoang, Akiko Eriguchi, and Huda Khayrallah. 2024.
\newblock X-alma: Plug \& play modules and adaptive rejection for quality translation at scale.
\newblock \emph{arXiv preprint arXiv:2410.03115}.

\bibitem[{Yang et~al.(2024)Yang, Yang, Zhang, Hui, Zheng, Yu, Li, Liu, Huang, Wei et~al.}]{yang2024qwen2}
An~Yang, Baosong Yang, Beichen Zhang, Binyuan Hui, Bo~Zheng, Bowen Yu, Chengyuan Li, Dayiheng Liu, Fei Huang, Haoran Wei, et~al. 2024.
\newblock Qwen2. 5 technical report.
\newblock \emph{arXiv preprint arXiv:2412.15115}.

\bibitem[{Yoo et~al.(2024{\natexlab{a}})Yoo, Park, Yun, Oh, and Lee}]{yoo2024code}
Haneul Yoo, Cheonbok Park, Sangdoo Yun, Alice Oh, and Hwaran Lee. 2024{\natexlab{a}}.
\newblock Code-switching curriculum learning for multilingual transfer in llms.
\newblock \emph{arXiv preprint arXiv:2411.02460}.

\bibitem[{Yoo et~al.(2024{\natexlab{b}})Yoo, Yang, and Lee}]{yoosafe}
Haneul Yoo, Yongjin Yang, and Hwaran Lee. 2024{\natexlab{b}}.
\newblock Code-switching red-teaming: Llm evaluation for safety and multilingual understanding.
\newblock \emph{arXiv preprint arXiv:2406.15481}.

\bibitem[{Yu et~al.(2023)Yu, Jiang, Shi, Yu, Liu, Zhang, Kwok, Li, Weller, and Liu}]{yu2023metamath}
Longhui Yu, Weisen Jiang, Han Shi, Jincheng Yu, Zhengying Liu, Yu~Zhang, James~T Kwok, Zhenguo Li, Adrian Weller, and Weiyang Liu. 2023.
\newblock Metamath: Bootstrap your own mathematical questions for large language models.
\newblock \emph{arXiv preprint arXiv:2309.12284}.

\bibitem[{Yu et~al.(2025)Yu, Dai, Wang, Wang, Chen, and Pei}]{yu2025opencsg}
Yijiong Yu, Ziyun Dai, Zekun Wang, Wei Wang, Ran Chen, and Ji~Pei. 2025.
\newblock Opencsg chinese corpus: A series of high-quality chinese datasets for llm training.
\newblock \emph{arXiv preprint arXiv:2501.08197}.

\bibitem[{Zellers et~al.(2019)Zellers, Holtzman, Bisk, Farhadi, and Choi}]{zellers-etal-2019-hellaswag}
Rowan Zellers, Ari Holtzman, Yonatan Bisk, Ali Farhadi, and Yejin Choi. 2019.
\newblock \href {https://doi.org/10.18653/v1/P19-1472} {{H}ella{S}wag: Can a machine really finish your sentence?}
\newblock In \emph{Proceedings of the 57th Annual Meeting of the Association for Computational Linguistics}, pages 4791--4800, Florence, Italy. Association for Computational Linguistics.

\bibitem[{Zhang et~al.(2023)Zhang, Aljunied, Gao, Chia, and Bing}]{zhang2023m3exam}
Wenxuan Zhang, Sharifah~Mahani Aljunied, Chang Gao, Yew~Ken Chia, and Lidong Bing. 2023.
\newblock \href {https://arxiv.org/abs/2306.05179} {M3exam: A multilingual, multimodal, multilevel benchmark for examining large language models}.

\bibitem[{Zhong et~al.(2024)Zhong, Cui, Guo, Liang, Lu, Wang, Saied, Chen, and Duan}]{zhong-etal-2024-agieval}
Wanjun Zhong, Ruixiang Cui, Yiduo Guo, Yaobo Liang, Shuai Lu, Yanlin Wang, Amin Saied, Weizhu Chen, and Nan Duan. 2024.
\newblock \href {https://doi.org/10.18653/v1/2024.findings-naacl.149} {{AGIE}val: A human-centric benchmark for evaluating foundation models}.
\newblock In \emph{Findings of the Association for Computational Linguistics: NAACL 2024}, pages 2299--2314, Mexico City, Mexico. Association for Computational Linguistics.

\bibitem[{Zhou et~al.(2024)Zhou, Wang, Huang, Huang, Han, Feng, Deng, Luo, and Chen}]{zhou2024moe}
Hao Zhou, Zhijun Wang, Shujian Huang, Xin Huang, Xue Han, Junlan Feng, Chao Deng, Weihua Luo, and Jiajun Chen. 2024.
\newblock Moe-lpr: Multilingual extension of large language models through mixture-of-experts with language priors routing.
\newblock \emph{arXiv preprint arXiv:2408.11396}.

\end{thebibliography}
